\documentclass{article}

\usepackage{arxiv}

\usepackage[utf8]{inputenc} 
\usepackage[T1]{fontenc}    
\usepackage{hyperref}      
\usepackage{url}            
\usepackage{booktabs}       
\usepackage{amsfonts}       
\usepackage{nicefrac}       
\usepackage{microtype}      
\usepackage{xcolor}         
\usepackage{graphicx}
\usepackage{multirow}
\usepackage{amsmath}
\usepackage{multicol}
\usepackage{float}
\usepackage{algorithmic}
\usepackage{algorithm}

\title{{SkyReels-Audio}: Omni Audio-Conditioned Talking Portraits in Video Diffusion Transformers}

\author{%
Zhengcong Fei, \ Di Qiu
\\
Kunlun Inc.\\
Beijing, China\\
{\tt\small \{author\}@gmail.com}
}

\begin{document}

\maketitle
\vspace{-0.5cm}
\begin{figure}[H]
  \includegraphics[width=\textwidth]{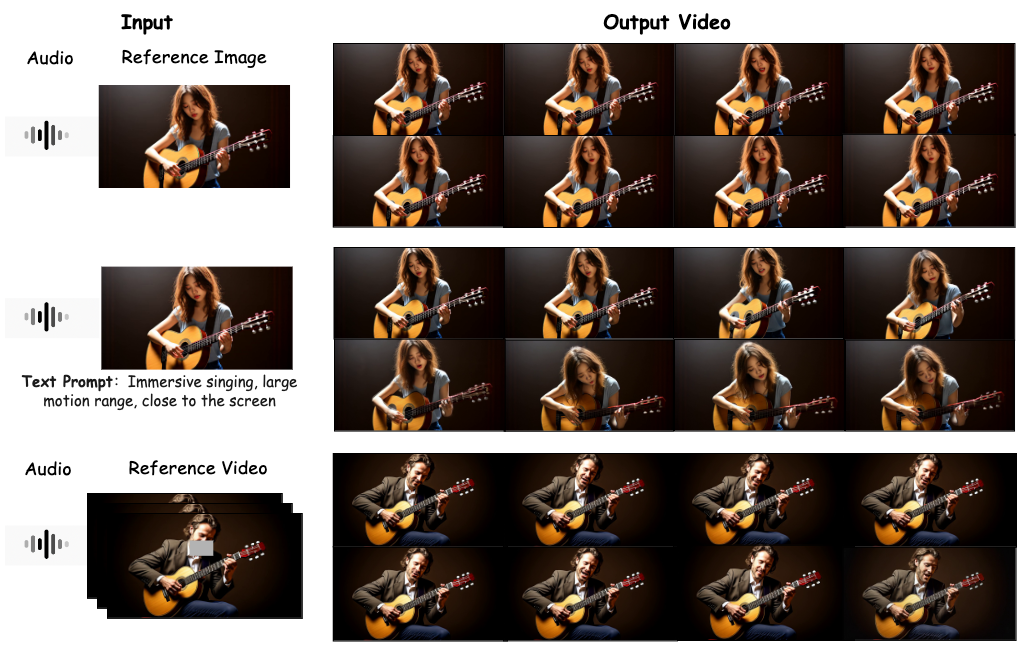}
  \vspace{-0.2cm}
  \caption{Given a portrait image, text, or video along with audio input, \texttt{SkyReels-Audio} can generate and edit portraits with strong identity consistency, expressive facial and natural body dynamics. In addition, \texttt{SkyReels-Audio} support infinite video generation based on various multi-modal controllable clues. 
  }
  \label{fig:case1} 
\end{figure}

\begin{abstract}
The generation and editing of audio-conditioned talking portraits guided by multimodal inputs, including text, images, and videos, remains under explored. In this paper, we present \texttt{SkyReels-Audio}, a unified framework for synthesizing high-fidelity and temporally coherent talking portrait videos. Built upon pretrained video diffusion transformers, our framework supports infinite-length generation and editing, while enabling diverse and controllable conditioning through multimodal inputs. 
We employ a hybrid curriculum learning strategy to progressively align audio with facial motion, enabling fine-grained multimodal control over long video sequences. To enhance local facial coherence, we introduce a facial mask loss and an audio-guided classifier-free guidance mechanism. A sliding-window denoising approach further fuses latent representations across temporal segments, ensuring visual fidelity and temporal consistency across extended durations and diverse identities. More importantly, we construct a dedicated data pipeline for curating high-quality triplets consisting of synchronized audio, video, and textual descriptions. Comprehensive benchmark evaluations show that \texttt{SkyReels-Audio} achieves superior performance in lip-sync accuracy, identity consistency, and realistic facial dynamics, particularly under complex and challenging conditions.
The model, along with demonstration videos, will soon be made publicly available at the project website: \url{https://www.skyreels.ai}.
\end{abstract}

\section{Introduction}

Recent developments in computer vision and human-computer interaction have highlighted the transformative potential of generating realistic digital avatars responsive to audio stimuli. These avatars, informed by static imagery, video sequences, or textual metadata, are poised to impact a wide range of domains, including digital storytelling, immersive education, and virtual communication platforms \cite{ma2021pixel, wang2021one, kal2024educational, rehm2016role, aicommunicate, johnson2018assessing,ji2021audio,lu2021live}. A foundational challenge in this domain lies in achieving precise, audio-conditioned modulation of avatar behavior, ensuring seamless integration with multimodal contextual signals. Robust synchronization across modalities is critical to enhancing user immersion and shaping future paradigms of interactive digital media consumption.

The field of talking head synthesis has witnessed significant progress through the application of advanced neural rendering techniques, such as GANs \cite{goodfellow2020generative}, NeRF \cite{mildenhall2021nerf}, and Gaussian Splatting methods~\cite{kerbl20233d}. 
These approaches~\cite{su2024audio, song2022audio, ye2024real3d, Chatziagapi2023, zhang2023sadtalker, ma2023styletalk, xing2023codetalker, bai2024efficient, peng2023emotalk, cho2024gaussiantalker, chen2024gstalker, li2024talkinggaussian, zhuang2024learn2talk, peng2024synctalk, li2024ae, liu2023moda, guan2023stylesync, wang2023seeing} have enabled the effective fusion of facial motion cues with static identity characteristics, producing highly realistic visual outputs. However, traditional parametric models \cite{tran2018_3DMM,li2017_FLAME} often fall short in capturing the full range of expressive motion, and conventional rendering methods continue to impose constraints on spatial resolution and output quality.
The advent of diffusion-based generative models~\cite{rombach2022LDM} has markedly improved the realism, diversity, and controllability of video synthesis~\cite{blattmann2023SVD, guo2023animatediff, xu2024easyanimate, wang2024magicvideo, yang2024cogvideox, zhou2024allegro, polyak2024movieGen, kong2024hunyuanvideo, wan2025, chen2025skyreels, fei2025ingredients, fei2025skyreels}. 
Notably, recent approaches~\cite{tian2024emo, wang2024V-Express, chen2024echomimic, yang2024megactor, jiang2024loopy, zheng2024memo, qiu2025skyreels} have demonstrated that diffusion priors are well-suited for disentangling spatial appearance from temporal dynamics, resulting in improved motion continuity.
Despite these advances, unified frameworks that integrate multimodal controls, particularly audio and text for talking head generation, remain underexplored. In particular, a comprehensive audio-conditioned diffusion model for robust long-range generation and editing across diverse portrait domains has yet to be fully investigated.

This paper introduces SkyReels-Audio, a omni audio-conditioned framework designed to generate and edit temporally consistent and visually realistic talking portrait videos using pretrained video diffusion transformers. 
The proposed system achieves high perceptual fidelity across a broad range of facial expressions and motion patterns, maintaining coherence even under diverse and dynamic conditions.
SkyReels-Audio facilitates portrait animation and editing by concurrently leveraging speech signals alongside text, images, and video inputs. In this multimodal configuration, the audio primarily governs articulatory movements, while auxiliary modalities influence expressive behaviors, physical gestures, and environmental transformations. 
We use Whisper \cite{radford2023robust} to encode audio information, which features are further aligned with video representations with audio 1D RoPE. 
These audio tokens are integrated into the denosing network via a cross-attention mechanism that allows effective fusion with video tokens.
To ensure robust audio control and multimodal integration, we employ a hybrid curriculum training strategy that incrementally conditions the model using joint optimization objectives. During inference, a bidirectional latent fusion algorithm is employed, enabling to produce indefinitely long video sequences while preserving temporal continuity and rendering quality across various speaker identities and settings.

Additionally, we perform a comprehensive evaluation using both a custom-designed benchmark and established public datasets. Our benchmark comprises over 50 audio-driven scenarios encompassing multiple languages and variable speaking styles and includes both quantitative metrics and human subjective assessments to capture perceptual and technical performance.
Under these settings, SkyReels-Audio demonstrates strong generalization capabilities across a diverse set of portrait images and video samples. The model effectively responds to audio-driven cues and textual instructions while preserving high visual quality and natural motion. Furthermore, it maintains consistent performance at varying sequence lengths, enabling efficient and scalable inference for real-world applications.
In summary, out contributions are listed as follows: 
\begin{itemize}
    \item We present SkyReels-Audio, a unified framework for audio-conditioned talking portrait generation and editing based on pretrained Video Diffusion Transformers. It employs a bidirectional latent fusion strategy to generate temporally coherent and visually consistent long-form videos across diverse speaking styles and contexts.
    \item We integrate a hybrid learning paradigm that combines image and video-based multimodal controls to improve model capacity and generalization. It introduces a facial region-weighted loss to balance local audio-driven articulation with global conditioning, enabling precise synthesis.
    \item We construct our approach on a carefully curated dataset of high-quality, temporally aligned audio-video-text triplets, supported by a comprehensive data pipeline, which facilitates effective multimodal learning.
    \item Experiments demonstrate that SkyReels-Audio achieves robust performance on diverse portrait inputs ranging from static images to dynamic videos under both speech and text conditioning, with efficient inference across variable sequence lengths.
\end{itemize}

\section{Method}

The goal of this work is to enable realistic portrait animation and editing guided by both speech audio and auxiliary multimodal inputs, including text, static imagery, or video references. The generated output is expected to retain the visual identity of the target subject as defined by the conditioning inputs, while faithfully capturing speech-related articulations such as lip synchronization, facial expressions, and head movements.


\subsection{Preliminaries}

\paragraph{Flow Matching.}
Flow Matching~\cite{lipman2022flow_match} provides a principled framework for mapping complex probability distributions to simpler ones by leveraging their probability density functions, thereby enabling sample generation via learned inverse transformations. 
Several recent works ~\cite{esser2024sd3,kong2024hunyuanvideo,chen2025skyreels,wan2025} serve as subclass of Flow Matching models that operate in the latent space, leveraging a pre-trained AutoEncoder~\cite{kingma2013VAE} to facilitate this process.
In contrast to conventional text-to-video models that rely exclusively on textual conditioning $T_s$, SkyReels-Audio incorporates multimodal conditioning signals, including the driving audio sequence ($A$), a static portrait image ($I_s$), a reference video clip ($V_s$), and the associated text prompt ($T_s$). Importantly, during training, any of the conditioning modalities may be randomly omitted to promote robust generalization and flexible inference. The model 
optimize to learn the reverse transformations with the objective,
\begin{equation}
\label{loss_func}
    L_{mse} =\mathbb{E}_{z_0,z_1,t \sim [0,1]} \bigg[ \Big\lVert {v_t}-{u_\theta} \big(z_t,t,T_s,I_s, V_s,A\big) \Big\lVert_2^2 \bigg],
\end{equation} 
where $u_\theta$ is a trainable denoising net. $z_1$ and $z_0$ notes the latent embedding of the training sample and the initialized noise sampled from the Gaussian distribution $\mathcal{N}(0,1)$. $z_t$ is the training sample constructed using a linear interpolation. Velocity $\mathbf{v}_t=d\mathbf{z}_t/dt=z_1-z_0$ serves as the regression target for the model.

\paragraph{Video Diffusion Transformers.}

The Diffusion Transformer represents a class of generative models built upon the Transformer architecture~\cite{peebles2023scalable}, exhibiting strong performance in video synthesis tasks by leveraging full spatio-temporal attention in three dimensions~\cite{kong2024hunyuanvideo, chen2025skyreels, wan2025}. In this work, we adopt SkyReels-V2~\cite{chen2025skyreels} as the core backbone. It incorporates a causal 3D VAE to perform joint compression across temporal and spatial axes, enabling efficient video encoding.
To follow textual guidance, it utilizes UMT5 to generate semantic embeddings from natural language inputs. These text features are then injected into the diffusion net via cross-attention layers, allowing conditioned video generation based on linguistic context. Additionally, temporal information is incorporated by predicting six distinct modulation parameters corresponding to each timestep, ensuring precise temporal alignment throughout the generative process.

\subsection{Model Architecture}

\begin{figure*}[t]
  \centering
   \includegraphics[width=1\linewidth]{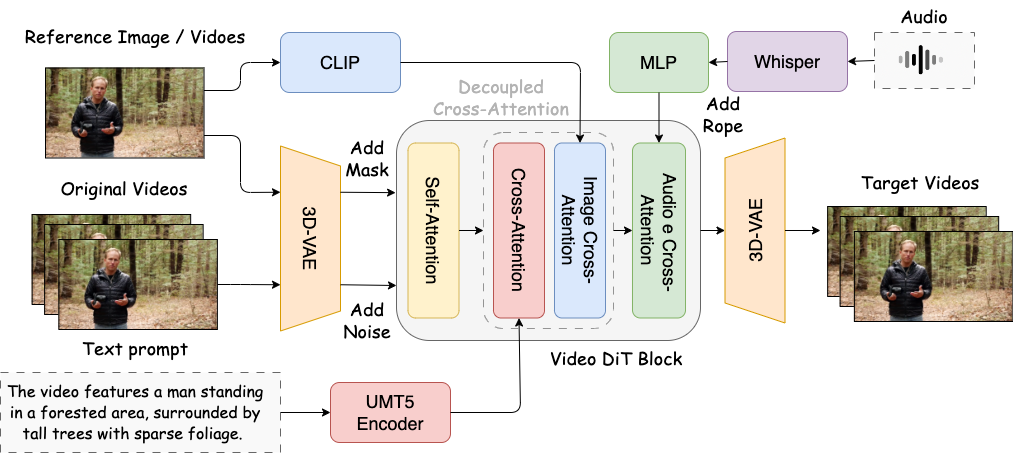}
   \caption{\textbf{Overview of SkyReels-Audio.} Whisper encodes resampled audio and fuse video tokens with cross-attention layers. Image and video controls are joint featured with VAE before combine with input noise to provide a video identity and environment priors.}
   \label{fig:framework} 
\end{figure*}

To enable unified control over both images and videos, we generalize the original image-to-video generation framework into a omni architecture, as illustrated in Figure \ref{fig:framework}. Specifically, a 3D VAE is employed to extract latent visual features, which are concatenated with a noise tensor along the channel dimension. To distinguish between static and dynamic inputs, we incorporate a binary temporal mask that encodes modality-specific information, allowing the model to differentiate between image and video sequences.
For conditioning on audio,  audio condition module processes the input wav signal with the Whisper feature extractor~\cite{radford2023robust}, which performs resampling and feature encoding. The resulting audio representations are passed through the Whisper encoder to obtain discrete token embeddings. These audio tokens are then injected into the video DiT using a dedicated cross-attention layers, 
which is strategically placed at the end of the decoupled cross attention blocks to modulate video generation based on the driving audio signal.

To improve the alignment between audio and visual modalities, we adopt RoPE ~\cite{su2024roformer}, which is particularly effective at capturing distance-aware relationships and generalizing to variable sequence lengths. Audio features are treated as one-dimensional sequences with shape 
$[1, L_{audio}]$, and the 1D RoPE is added accordingly in attention operation. This technique enhances both intra-modal coherence and cross-modal correspondence, contributing to more accurate lip synchronization and improved semantic consistency in generated visual content.

\subsection{Hybrid Learning Strategy}


For interleaved image animation or video editing tasks, we adopt a hybrid learning strategy that further improves audio-motion alignment.
Experiments revealed that when employing the joint training strategy, even with a T2V model as the base model, satisfactory image animation results could still be obtained. In contrast, training the image animation task alone often required longer convergence times and sometimes failed to produce correct results. Furthermore, in the image animation task, the control signals are limited to text, images, and audio. Given the strong correlation between audio and lip movements, the model more readily learns audio-driven synthesis. 
Conversely, in the video editing task, the inclusion of additional video control conditions introduces dependencies between lip movements and surrounding regions. The presence of these peripheral motions tends to diminish the influence of audio control. 

Therefore, we employ masks to differentiate between the Image Animation and Video Editing tasks. For the Image Animation task, the input image serves as the reference frame and is concatenated with a sequence of empty frames to form the video input $V_{s}=(I_s, I_{empty},...,I_{empty})$ for the 3D VAE. The corresponding mask sequence $V_M=(M_{ones}, M_{zeros},...,M_{zeros})$ undergoes max pooling with the same downsampling factor as the 3D VAE and is then concatenated with the VAE output along the channel dimension. In the Video Editing task, the first frame of the input video is treated as the reference frame. For the remaining frames, we detect facial landmarks using DWPose and generate lower-face bounding boxes. These frames are then masked based on the BBox to produce the corresponding video and mask sequences $V_s=(I_0, I_1^{mask}, ..., I_n^{mask})$, $V_m=(M_{ones},M_{mouth},...,M_{mouth})$. Subsequently, the same processing pipeline as in Image Animation is applied.
To prioritize the generation quality within the masked regions, we adapt the flow matching loss function by applying distinct weighting factors to the masked and non-masked areas as:
\begin{equation}
    L_{joint} = w_1 * V_m^{downsample} * L_{mse} + w_2 * (1 - V_m^{downsample}) * L_{msk}
\end{equation}

To guide the model's focus toward the articulation-relevant regions, particularly the lips, we leverage DWPose \cite{yang2023effective} to extract precise masks in pixel space, which are subsequently transformed into the latent representation space through trilinear interpolation, yielding the constraint mask denoted as $\mathcal{M}_{lip}$. To maintain a balance between local precision in lip synchronization and the preservation of global visual realism, we use a probabilistic gating mechanism governed by a threshold $p_{mask}$ as:
\[
L_{face} =
\begin{cases}
L_{mse}, &\quad\text{if } p\ge p_{mask}\\
\mathcal{M}_{lip}   \odot  L_{mse} &\quad\text{otherwise} \\
\end{cases}
\]
where $\odot$ denotes element-wise multiplication. This adaptive masking strategy ensures that the model places selective emphasis on the lip region during training while retaining the overall structural integrity of the generated portrait.

\subsection{Inference Optimization} 


\paragraph{Audio CFG.}
Prior research has highlighted the limitations of text-to-image generation models in accurately adhering to input textual descriptions~\cite{esser2024sd3,sid_lsg}. To address such shortcomings, these approaches often employ inference-time guidance strategies to improve alignment between generated content and conditioning inputs. Motivated by these insights, we propose an Audio-Guided Conditional Sampling mechanism to enhance synchronization with driving audio signals during inference.
The adjusted denoising function incorporating both audio and text guidance is formulated as:
\begin{equation}
\begin{split}
    & \hat{u_\theta}^\text{cfg} = (1+\omega_\text{audio}){u_\theta}(\mathbf{z}_t, t, T_s, I_s, V_s, A) - \omega_\text{audio}{u_\theta}(\mathbf{z}_t, t, T_s, I_s, V_s, \emptyset ) \\
    &\quad\quad\quad\quad+(1+\omega_\text{text}){u_\theta}(\mathbf{z}_t, t, T_s, I_s, V_s, \emptyset ) - \omega_\textrm{text}{u_\theta}(\mathbf{z}_t, t, \emptyset, \emptyset, \emptyset, \emptyset ), \\
    \label{eq:cfg}
\end{split}
\end{equation}
where $\omega_\text{audio}$ and $\omega_\text{text}$ represent the CFG scales specifically designed for audio condition and text condition, respectively.  Note that we adopt time-dependent scheduling for these CFG weights, allowing the model to dynamically balance conditioning influences across the diffusion trajectory, thereby improving fidelity and robustness in audio-synchronized portrait generation.

\begin{figure}[t]
  \centering
  \includegraphics[width=0.8\textwidth]{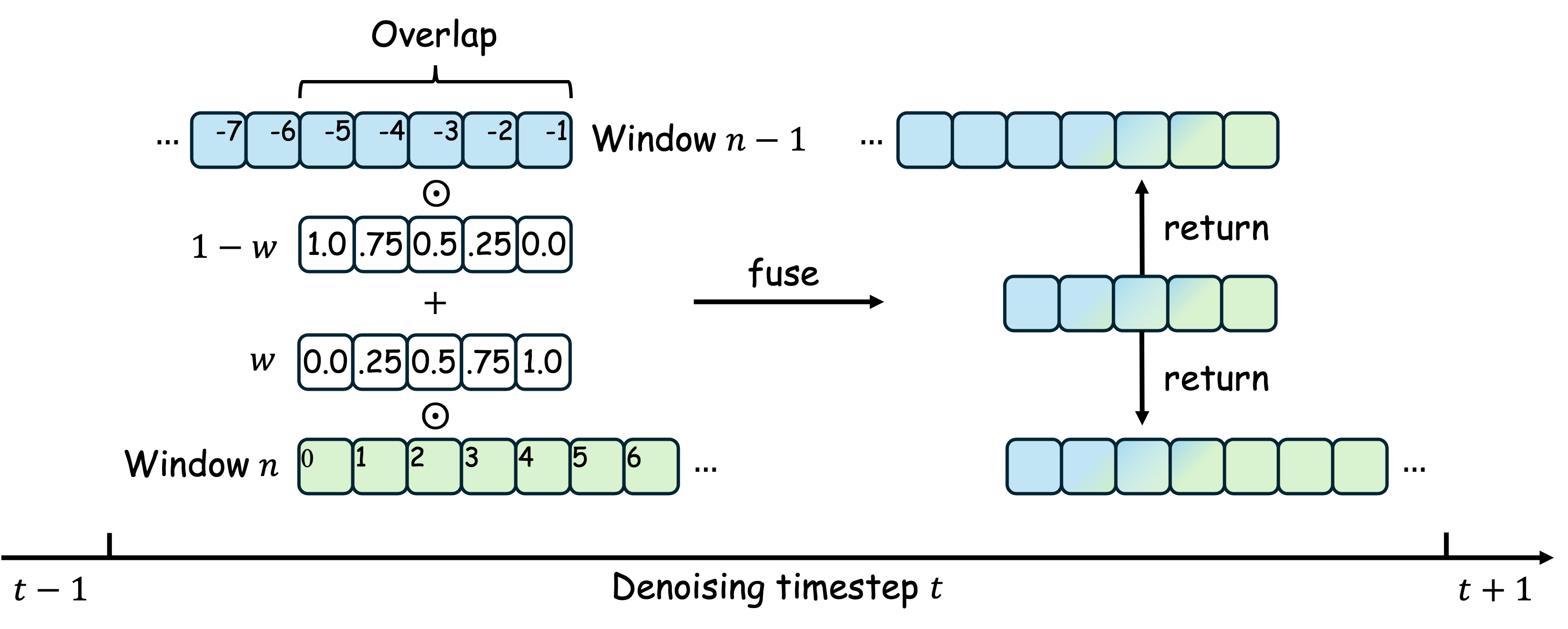}
  \caption{\textbf{Illustration of BLF.} BLF is a tuning-free overlapping sliding window strategy, performing bidirectional fusion of the latents within adjacent windows in the same denoising step.
  }
  \label{fig:BLF} 
\end{figure}

\paragraph{Infinite Video Generation via Bidirectional Latent Fusion.}

We introduce a tuning-free infinite video inference strategy, named as Bidirectional Latent Fusion (BLF). During denoising loop, BLF achieves smooth transitions between different video windows by bidirectionally and weightedly fusing video latents. 
Unlike the motion frames based methods \cite{tian2024emo, jiang2024loopy, cui2024hallo3}, our BLF requires no training support and significantly reduces image quality degradation caused by error accumulation. Compared to non-overlap methods \cite{ji2024sonic}, our approach provides greater image stability.

Specifically, as illustrated in Figure \ref{fig:BLF}, BLF comprises three key phases: (i) We resample audio seuqences to ensure temporal alignment with video frames, while simultaneously initializing a noise vector of corresponding duration. (ii) Within the same denoising step, overlapping latents between adjacent windows is weighted fused through a sliding window mechanism, where the fusion weights are linearly interpolated based on the relative frame indices. (iii) The fused latents are reinserted into both participating windows, thereby ensuring that both ends of the middle window are fusion features. This approach effectively achieves smooth transition between different windows through bidirectional feature propagation. The total process is listed in Algorithm \ref{alg:BLF}.
Note that we found that the color of the images has a probability of becoming darker during long video inference process, and this phenomenon has also been observed in other DiT works. We mitigate this issue by strictly controlling the quality of the training data and performing color-unification post-processing on the generated videos.


\begin{algorithm}
    \caption{Algorithm of BLF}
    \label{alg:BLF}
    
    \begin{algorithmic}[1]
        \REQUIRE Denoising steps $T$, audio embedding $c^{[0,l]}_{a}$ and initial noisy latent $z^{[0,l]}_{T}$ with length $l$, pretrained DiT model $\text{DiT}(\text{·})$ for sequence length $f$, $f \le l$, window size consistent with $f$, overlap length $o$ between every two windows.
        \ENSURE Denoised latent $z^{[0,l]}_{0}$.
        
        
        \FOR{$t = T,...,1$}
            \STATE Initialize start index $s = 0$, end index $e = s + f$, previous end index $e_{prev} = e$.
            \WHILE{$e \le l$} 
            \STATE $z^{[s,e]}_{t-1} = \text{DiT}(z^{[s,e]}_{t}, c^{[s,e]}_{a}, t)$
            \IF {$s \ne 0$ and $t \ne T$}
                \STATE $w = \text{zeros}(o)$
                \FOR{$i = 1,...,o$}
                \STATE $w_i = \frac{i-1}{o-1}$
                \ENDFOR
                \STATE $z^{[s,s+o]}_{t-1} = w * z^{[s,s+o]}_{t-1} + (w - 1) * z^{[e_{prev}-o,e_{prev}]}_{t-1}$
            \ENDIF
            \IF {$e \ne l$} 
            \STATE $e_{prev} \gets e, s \gets  s + (f - o), 
                    e \gets \left\{\begin{array}{ll}
                        s+f & \text { if } s+f<l \\
                        l & \text { otherwise }
                        \end{array}\right.$
            \ENDIF
    		\ENDWHILE 
        \ENDFOR
        
        \RETURN Denoised latent $z^{[0,l]}_{0}$
    \end{algorithmic}
\end{algorithm}

\paragraph{Hybrid Inference Strategy}
Benefiting from the joint training of image animation and video editing tasks, our model supports both image and video inputs during inference.  Experimental results demonstrate that when driven by the same audio input, the video generated from a single image (i.e., the first frame of a video) exhibits superior lip-sync accuracy compared to videos generated from full video inputs.
To enhance audio-visual synchronization in video editing tasks, we propose a hybrid inference strategy. \textbf{Early Denoising Steps (First N steps)}: Use the full video input to maintain structural consistency with the source video. \textbf{Later Denoising Steps (Remaining steps)}: Switch to image input (first frame only) to refine lip-sync details, while adaptively adjusting the corresponding mask sequence.
\begin{equation}
u_\theta^t =
    \begin{cases}
    {u_\theta}(\mathbf{z}_t, t, T_s, V_s^V, V_M^V, A), &\quad\text{if } t\le N\\
    {u_\theta}(\mathbf{z}_t, t, T_s, V_s^I, V_M^I, A), &\quad\text{otherwise} \\
    \end{cases}
\end{equation}

\paragraph{Model Acceleration.}
To accelerate the inference process, we implemented the two major optimizations. We employed Teacache~\cite{liu2024timestep} to eliminate redundant denoising steps through latents reuse. To reduce the consumption of VRAM, the computation of Teacache is transferred to the CPU, and the increased inference time brought about by this process can be ignored. In addition, we adopted Unified Sequence Parallelism (USP)~\cite{fang2024unified} to enable multi-GPU inference. With the increase in the number of nodes, in order to ensure that the data shapes split to each node are the same, some additional cropping of the input reference iamge is inevitable. 
Notably, TeaCache and USP can be activated simultaneously. Consequently, our framework achieves generating 80 frames of video within a minute (conducted 50 inference steps on 8 A800 GPUs) while incurring no perceptible quality degradation. Quantitative analysis of acceleration performance will be presented in the experiments section.

\subsection{Data Pipeline}

\begin{figure}[H]
  \centering
  \includegraphics[width=\textwidth]{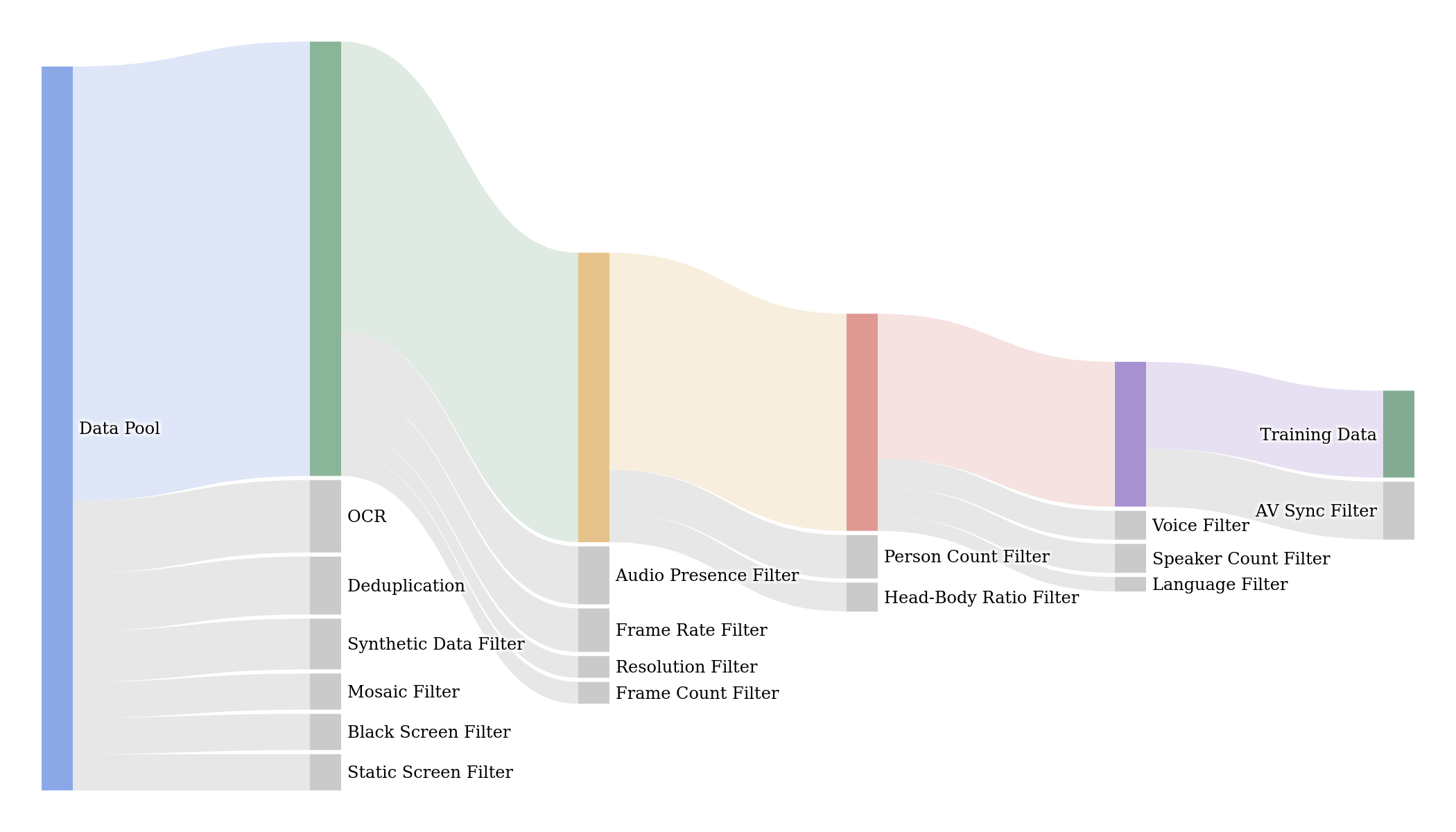}
  \caption{\textbf{Data Processing Pipeline.} This is a data funnel to filter high-quality video data.
  }
  \label{fig:data_pipeline} 
\end{figure}

To enhance the quality of model training, we constructed a data processing pipeline as shown in Figure \ref{fig:data_pipeline}, a progressive filtering strategy adopted to strictly monitor the training data. 
Specifically, we collected 10K hours of video data from public datasets (including OpenHumanVid~\cite{li2024openhumanvid}, Panda-6M~\cite{wang2020panda}, Hallo3~\cite{cui2024hallo3}) and self-collected sources, placing it into a raw Data Pool. Then, we processed the data in stages based on image content, video quality, portrait quality, audio quality, and audio-visual synchrony, ultimately obtaining 1K hours for training. At the same time, we have introduced manual annotation in the data processing flow of each stage to ensure that the bad case rate remains below 5\%.


Our data preprocessing pipeline begins with the collection of a large-scale video dataset, which is segmented into short clips based on content coherence. We apply our video captioning model, SkyCaptioner-V1~\cite{chen2025skyreels}, to generate descriptive annotations for each clip, providing high-quality text supervision. To analyze human presence and interaction, we use YOLO-World~\cite{cheng2024yolo} and InsightFace for body and face detection, respectively, allowing us to estimate the number of individuals in each clip. We extract pose-related features using DWpose to compute head-to-body ratios, and apply Whisper to identify the spoken language. To evaluate audiovisual synchronization, we compute the sync confidence score~\cite{chung2017syncNet}, and employ a source separation model to estimate the number of distinct speakers. This multi-stage preprocessing ensures rich, multimodal supervision for downstream learning tasks.


\section{Experiments}

\subsection{Experimental Settings}

\paragraph{Implementation Details}

We train the model using internally collected data constructed through the data pipeline process in section 2.5, which results in approximately 1K hours dataset. To construct a coarse-to-fine dataset, we apply a filtering mechanism to the audio component, leveraging synchronization and offset criteria. 
We train the SkyReels-Audio based on SkyReels-V2 backbone \cite{chen2025skyreels}. 
Training is carried out in two distinct stages. During the initial phase, the model is trained solely on audio data to establish robust foundational alignment between the audio and visual modalities. In the subsequent phase, we adopt a joint training strategy that incorporates both image and video inputs, aimed at improving motion consistency and temporal coherence in the generated outputs.
We used AdamW \cite{diederik2014adam} optimizer with a constant learning rate of $10^{-5}$ for all trainable modules across all stages. We employ Flow matching to train the model, with the entire training conducted on 80 Nvidia GPUs. 
To enhance video generatiob varibility, the reference image / video, guiding audio, prompt are each seto to be independently discarded with a probability of 0.15. During inference, we employ the sampling steps of 50, the audio CFG is set to 4.5.

\paragraph{Evaluation Metrics.}

We employ Q-align \cite{wu2023q} visual language model to evaluate video quality (IQA) and aesthetic metrics (ASE), and use FID \cite{heusel2017fid_metric} and FVD \cite{unterthiner2019fvd_metric} to assess the distance between generated videos and real videos. Finally, we utilize Sync-C and Sync-D as proposed in SyncNet~\cite{chung2017syncNet} for audio-visual synchronization estimation.

\subsection{Main Results}

\begin{table}[t]
\begin{center}
\setlength{\tabcolsep}{3.5mm}
\renewcommand{\arraystretch}{1.1}
\begin{tabular}{lcccccc}
\toprule 
 & \multicolumn{6}{c}{HDTF} \\
  Method & FID $\downarrow$ & FVD $\downarrow$ & Sync-C  $\uparrow$ & Sync-D $\downarrow$   & IQA  $\uparrow$  & ASE  $\uparrow$  \\
\midrule
Hallo3\cite{cui2024hallo3}  & 40.12& 408.12 & 5.75& 10.12 &4.03& 2.65  \\
FantacyTalking\cite{wang2025fantasytalking} & 39.53 &381.22 & 5.36  & 11.68 &3.50 & 2.38 \\
SkyReels-Audio & 38.32 & 364.71 & 6.06 &9.12 & 4.60 &2.92 \\
\bottomrule
\end{tabular}
\end{center}
\caption{Quantitative comparison of audio-driven synchronization metrics on HDTF dataset, benchmarked against open-source models.}
\label{tab:quantitative_combined1}
\end{table}

\begin{table}[t]
\begin{center}
\setlength{\tabcolsep}{2.0mm}
\renewcommand{\arraystretch}{1.1}
\begin{tabular}{lcccccc}
\toprule 
 & \multicolumn{4}{c}{HDTF} & \multicolumn{2}{c}{User Study(Internal)}\\
  Method & FID $\downarrow$ & FVD $\downarrow$ & Sync-C  $\uparrow$ & IQA  $\uparrow$ & AV Consist. $\uparrow$ & Visual Quality $\uparrow$ \\
\midrule
LatenSync\cite{li2024latentsync}   & 40.23 & 390.25 &8.48  & 3.63  & 1.19 & 1.00\\
SkyReels-Audio &39.75 &377.23 &8.49  &3.62  & 1.38 & 1.32\\
\bottomrule
\end{tabular}
\end{center}
\caption{Quantitative comparison of automatic metrics on HDTF dataset for Lip Sync task.}
\label{tab:edit}
\end{table}

\begin{table}[t]
\begin{center}
\setlength{\tabcolsep}{5.5mm}
\renewcommand{\arraystretch}{1.1}
\begin{tabular}{lcccc}
\toprule 
 & \multicolumn{4}{c}{Internal} \\
  Method & Sync-C  $\uparrow$ & Sync-D $\downarrow$   & IQA  $\uparrow$  & ASE  $\uparrow$  \\
\midrule
Hallo3\cite{cui2024hallo3} & 5.53 & 9.33 & 4.13 & 2.80 \\
MagicInfinite\cite{yi2025magicinfinite} &6.22 & 8.43 & 4.56 & 3.00  \\
OmniHuman-1\cite{ominihuman} &7.50 & 7.47 & 4.66 & 3.19  \\
FantacyTalking\cite{wang2025fantasytalking} & 3.67 & 10.97 & 4.26 & 2.80 \\
SkyReels-Audio &6.75 & 8.32 & 4.42 & 2.91 \\
\midrule
Audio CFG=1 &5.78 & 9.65 & 4.58 & 3.02  \\
Audio CFG=3 &6.30 & 8.78 & 4.45 & 2.91 \\
w/o Audio RoPE & 5.58 & 9.75 & 4.35 & 2.82  \\
\bottomrule
\end{tabular}
\end{center}
\caption{Quantitative comparison of audio-driven synchronization metrics on internal dataset. We also ablate the Audio CFG and Audio RoPE to illustrate the effectiveness of design.}
\label{tab:quantitative_combined2}
\end{table}

\begin{figure*}
  \centering
   \includegraphics[width=1\linewidth]{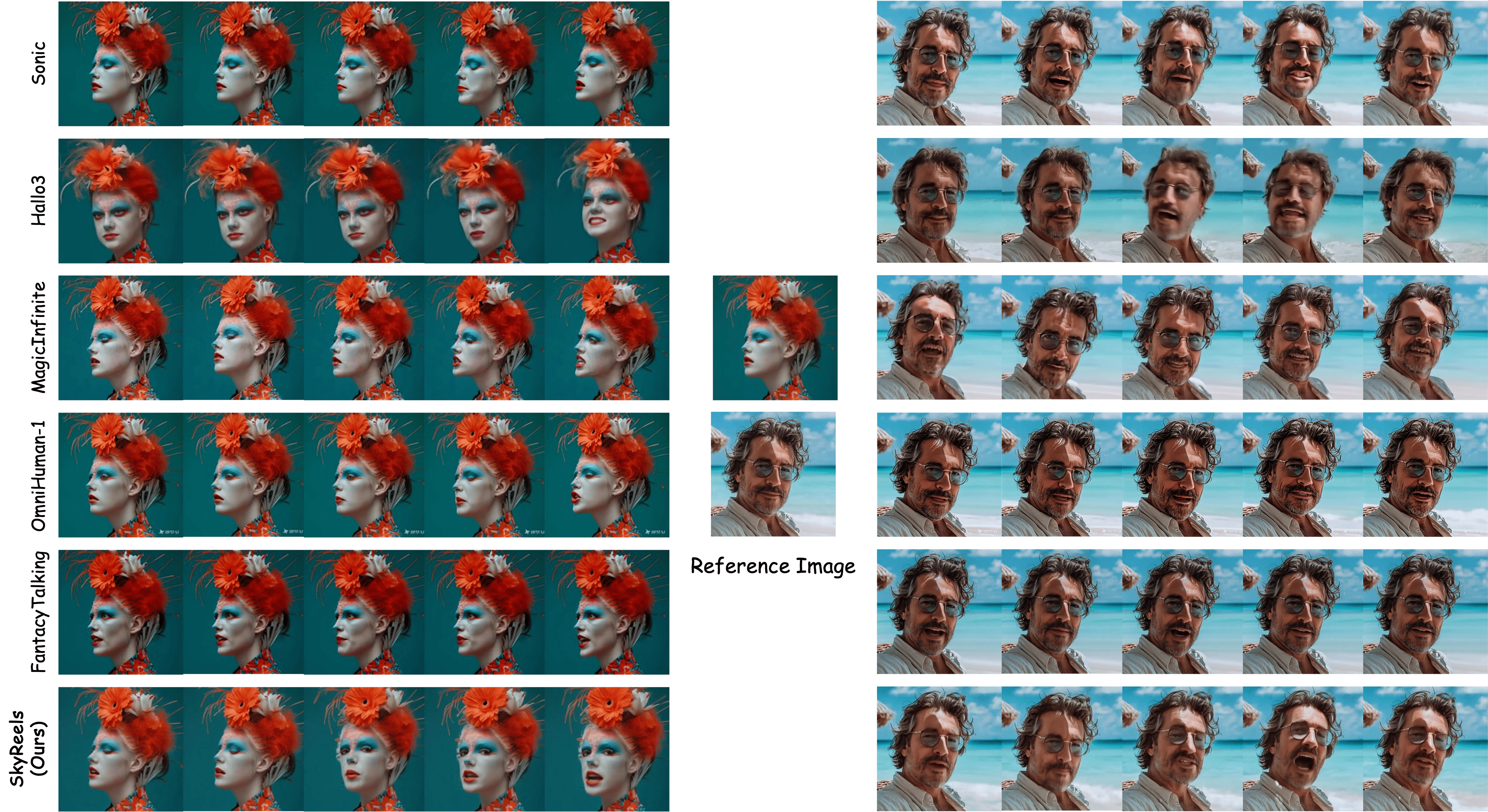}
   \caption{\textbf{Qualitative comparisons with other audio-driven talking portrait methods.} Our approach produce more accurate lip synchronization with naturalness. }
   \label{fig:results_sota} 
\end{figure*}

\begin{figure*}
  \centering
   \includegraphics[width=1\linewidth]{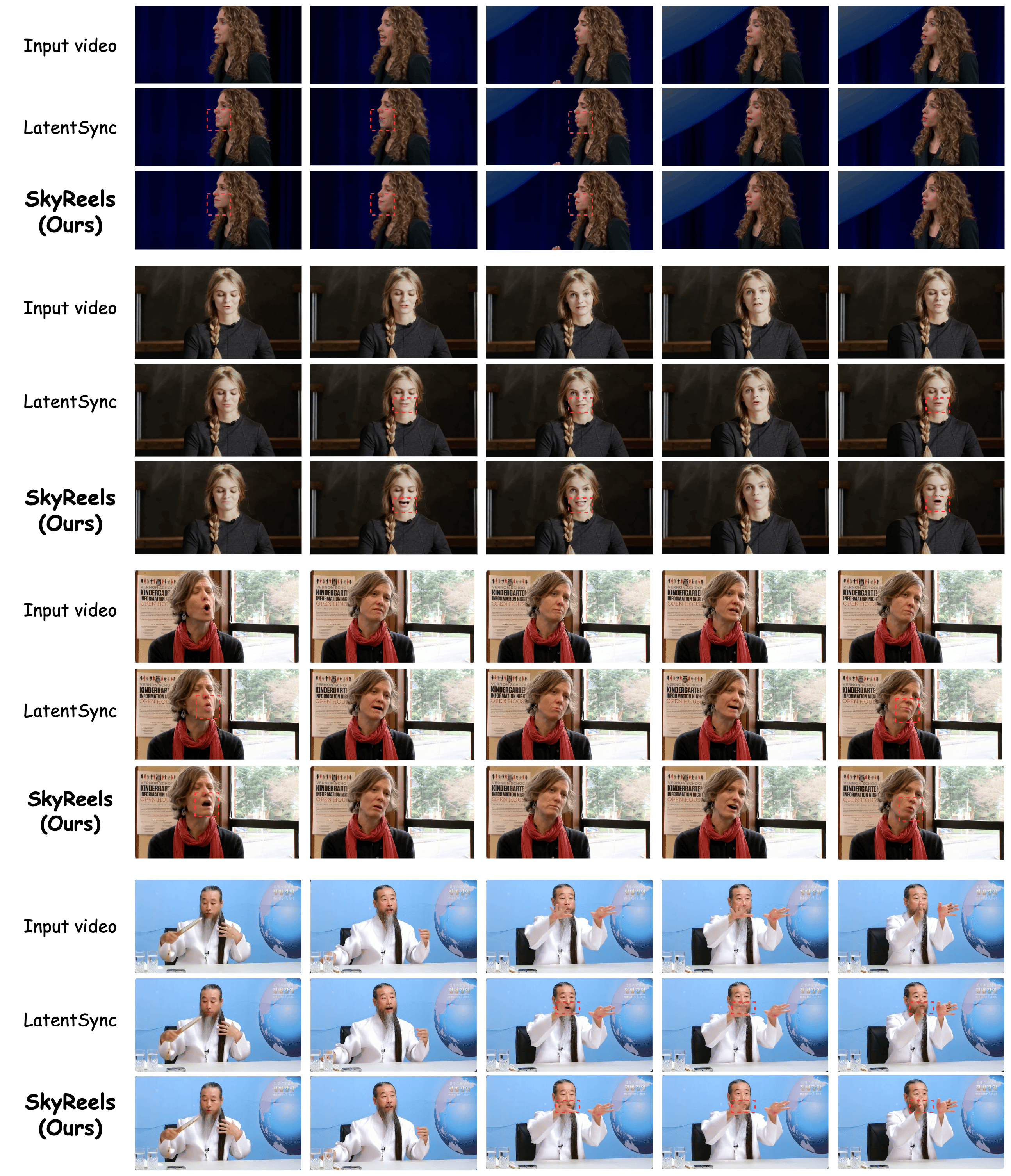}
   \caption{\textbf{Qualitative comparisons with SoTA lip-sync methods.} We can see that SkyReels-Audio create better audio lip alignment compared with the baseline. }
   \label{fig:results_sota_retalking} 
\end{figure*}

\begin{figure*}
  \centering
   \includegraphics[width=1\linewidth]{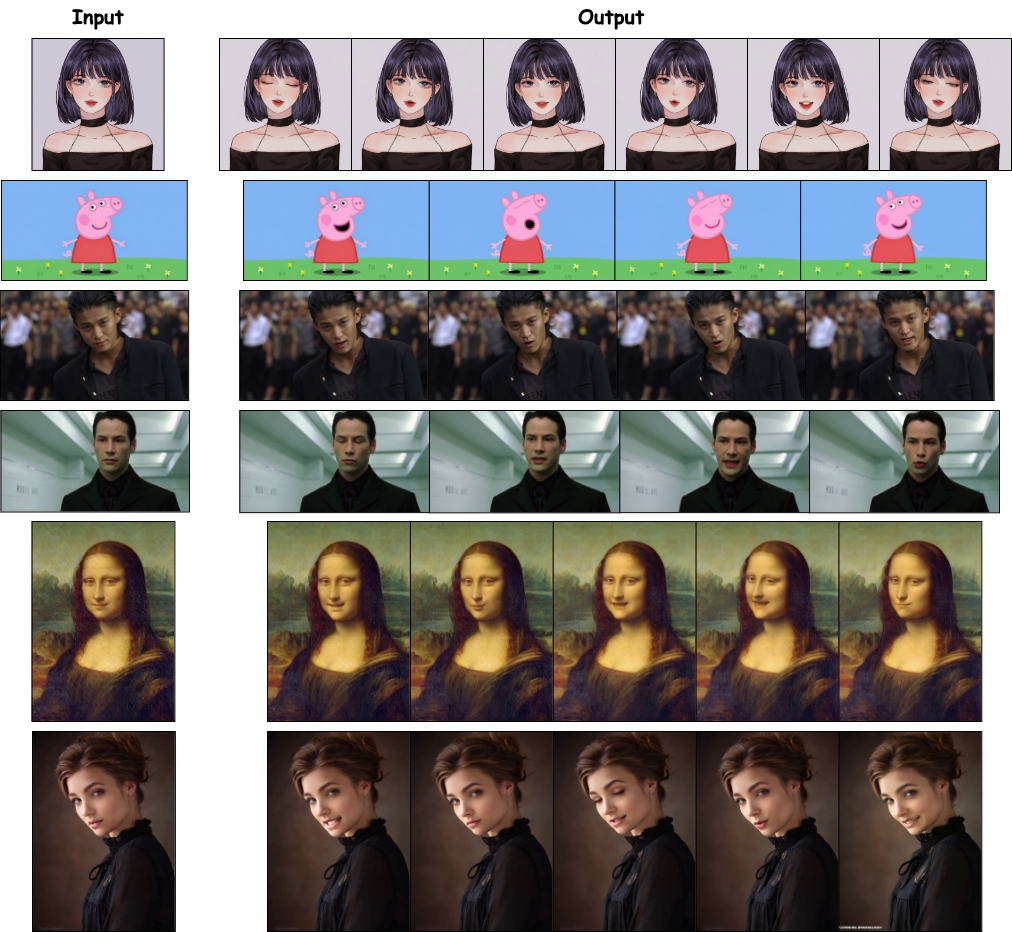}
   \caption{\textbf{More generated results of \texttt{SkyReels-Audio}.} Our approach can handle reference images of different objectives, sizes, and styles, and claim naturally consistent video results. }
   \label{fig:results_more} 
\end{figure*}

\paragraph{Qualitative Analysis.}
To comprehensively evaluate the capability of the proposed model in animating any audio-driven portrait styles and its generalizability across diverse application contexts, we present a newly curated benchmark dataset. This benchmark comprises 50+ portrait images spanning multiple domains, including stylistic variations such as anime, sculpture, and photorealistic renderings, all synthesized using advanced text-to-image generation techniques.
In addition, the benchmark incorporates 30 distinct audio segments reflecting a range of language, vocal scenarios, including singing, spoken word, and rap performances. Complementary to these are 20 textual prompts designed to elicit specific emotional states and physical gestures during speech. Certain prompts also contain descriptions of dynamic environmental contexts tailored to particular portrait backgrounds—for instance, the movement of foliage in the wind or the sound of ocean waves.
This comprehensive benchmark, characterized by its stylistic breadth and contextual richness, is intended to support broader research efforts within the community with public.
We assess the performance of SkyReels-Audio using this internal benchmark. As illustrated in Figure \ref{fig:results_sota}, \ref{fig:results_sota_retalking} and \ref{fig:results_more}, the model achieves high perceptual fidelity and temporal coherence, even under conditions of computational acceleration. Notably, despite being trained solely on real-world portrait videos, SkyReels-Audio demonstrates strong generalization to stylized portraits, indicating robust cross-domain adaptability.

\paragraph{Quantitative Analysis.}


To assess the fidelity of generated portrait videos across various methods, we employ standardized evaluation metrics. Our evaluation is conducted on a test set comprising 100 video clips randomly sampled from the HDTF dataset~\cite{zhang2021hdtf} and the internal dataset, both of which were excluded from the model’s training data.
For each test instance, the initial video frame is used as a static portrait reference, while the corresponding audio track drives the generation of a full video sequence. The original video clip serves as the ground truth. Additionally, descriptive textual prompts are extracted from each test video using SkyCaptioner-V2 to support multimodal conditioning.
As reported in Table \ref{tab:quantitative_combined1}, \ref{tab:edit}, and \ref{tab:quantitative_combined2}, SkyReels-Audio consistently outperforms baseline models in terms of visual fidelity, motion realism, and lip-sync precision, achieving comparable with close-source models. Notably, the model maintains competitive performance while achieving a significant inference acceleration. 

\paragraph{User Study.}

To further validate the effectiveness of our method, we conducted a subjective evaluation on the Internal dataset. Specifically, each participant assessed two key dimensions: audio-visual consistency (AV Consist.) and visual quality. A total of 20 participants rated each aspect on a scale for 0 to 2 (from bad to good). As shown in Table \ref{tab:edit}, the results indicates that SkyReels-Audio outperform baselines in both evaluation dimensions.

\subsection{Ablation Study}

\paragraph{Audio CFG.}

We first analyze the impact of Audio CFG in Equation \ref{eq:cfg}. 
Specifically, we set different Audio CFG values range from [1, 3, 4.5,], while keep other inference parameters same, and perform inference on the Internal dataset. 
The results are listed in Table \ref{tab:quantitative_combined2}.
We can see that, as the Audio CFG value increases, the metrics related to audio-visual consistency, i.e., Syn-C and Syn-D, continue to improve, but the video visual quality will decrease slightly. Taking into account both factors, we set a Audio CFG value of 4.5 by default.

\paragraph{Audio RoPE.}

We then verify the effect of Audio RoPE incorporation when fuse video and audio tokens in cross-attention layers.  
Table \ref{tab:quantitative_combined2} shows the evaluated result with or without audio position encoding. It can be clearly observed that the introduction of position encoding effectively improves the alignment between visual quality and audio, helping the model to more accurately locate useful information.

\begin{figure}[H]
  \centering
  \includegraphics[width=\textwidth]{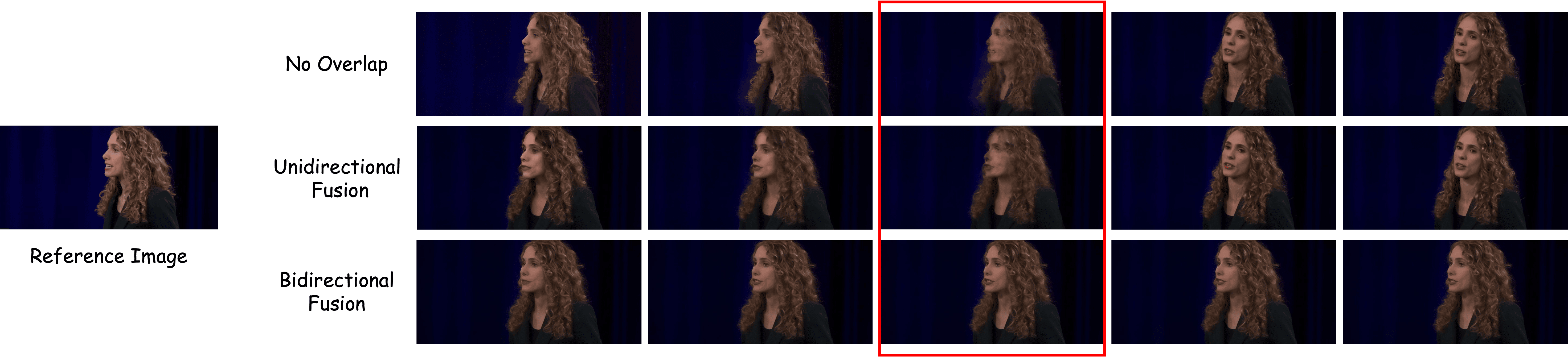}
  \caption{Qualitative comparison of different latent fusion methods. BLF generates long videos with less video ghosting and more coherent movements.}
  \label{fig:effect_of_BLF} 
\end{figure}

\paragraph{Effect of BLF.}
Figure \ref{fig:effect_of_BLF} provides a qualitative comparison of different latent fusion methods. The results show that when no overlap is set between sliding windows, there is a significant jump in the image, because each video clip is generated independently based on the reference image; when only unidirectional latent fusion is used, there is still an discontinuity at the stitching area; our bidirectional fusion method significantly improves the stability of the transition frames.

\paragraph{Effects of TeaCache and USP.}
Table \ref{tab:teacache&usp} illustrates the effects of TeaCache and USP on inference time. The experiment was on A800 GPU, with the inference steps set to 50. We tested the average inference time of 50 cases with 80 frames, and with both TeaCache with threshold $\alpha$=0.3 and USP turned on, the inference speed increases by about 24 times (from 23.62 minutes to 0.97 minute), while the model performance does not show a significant drop.

\begin{table}[]
\begin{center}
\begin{tabular}{@{}lcccc@{}}
\toprule
             & 1 GPU (w/o USP) & 2 GPUs (USP) & 4 GPUs (USP) & 8 GPUs (USP) \\ \midrule
w/o TeaCache & 23.62 min       & 17.90 min    & 7.80 min     & 4.29 min     \\
w/ TeaCache ($\alpha$ = 0.3)     & 8.96 min        & 7.18 min     & 1.88 min     & 0.97 min     \\ \bottomrule
\end{tabular}
\end{center}
\caption{The effect of TeaCache and USP on inference time. Both methods can effectively improve the inference speed of video generation in our method.}
\label{tab:teacache&usp}
\end{table}

\section{Related Work}
\subsection{Diffusion-based Lip Sync}

Generating photorealistic talking-head videos conditioned on audio input continues to pose significant challenges within the domain of multimodal synthesis. 
Earlier approaches~\cite{zhang2023sadtalker, ma2023dreamtalk, wei2024aniportrait} predominantly leveraged 3D intermediate representations, where facial animation parameters derived from 3DMMs were used to guide video synthesis. However, such pipelines often fall short in capturing the subtle intricacies of facial expressions and head dynamics, limiting the realism and emotional expressiveness of the generated content. 
To overcome these constraints, recent efforts have shifted towards fully end-to-end diffusion-based frameworks \cite{fei2024diffusion,fei2024scalable} for audio-driven lip sync animation~\cite{tian2024emo, jiang2024loopy, chen2024echomimic, cui2024hallo3}, which show increased promise. 
LatentSync \cite{li2024latentsync} use SyncNet loss to help better audio-lip alignment in latent diffusion framework. 
EMO~\cite{tian2024emo} and V-Express~\cite{wang2024V-Express} have leveraged audio inputs to drive precise lip synchronization while incorporating sparse visual cues to animate head dynamics, resulting in compelling audiovisual coherence.
However, two key challenges remain unresolved. First, identity preservation is typically handled by reference encoders adapted from general-purpose vision backbones, which restricts the capacity to generate complex and diverse motion patterns. Second, current systems still struggle to capture the broader spectrum of expressive behaviors—such as micro-expressions, facial gestures, and upper-body movements—that are only weakly correlated with the audio signal.
Our model also build on advanced diffusion model, while integrate multi-modal controls and joint modeling for better lip sync alignment.

\subsection{Diffusion-based Portrait Animation}
Diffusion models\cite{rombach2022LDM} have emerged as a cornerstone in the field of generative media synthesis, demonstrating remarkable efficacy in producing both images and videos \cite{blattmann2023SVD, guo2023animatediff, chen2023pixart, xu2024easyanimate, wang2024magicvideo, qiu2024moviecharacter,fei2025ingredients}.
Within the domain of portrait animation, Echomimic~\cite{chen2024echomimic} and MegActor-Sigma~\cite{yang2024megactor}—enhance controllability over animation by jointly modeling visual and auditory signals. Concurrently, methods like \cite{jiang2024loopy, zheng2024memo, fang2024motioncharacter} prioritize temporal consistency and affective realism, introducing emotion-aware modules and frame-level blending strategies specifically designed for sustained video generation. Despite these advancements, existing pipelines often struggle with preserving visual quality and narrative flow over time, primarily due to the compartmentalized treatment of spatial and temporal dependencies via isolated attention mechanisms.
In contrast, recent DiT based architecture—such as OminiHuman-1~\cite{ominihuman}, CogVideoX~\cite{yang2024cogvideox}, Allegro~\cite{zhou2024allegro}, MovieGen~\cite{polyak2024movieGen}, and HunyuanVideo~\cite{kong2024hunyuanvideo}—employ integrated 3D full-attention mechanisms that offer unified modeling of space-time information, thereby producing higher-quality video outputs. Drawing on these insights, we also resort it into our portrait animation framework, achieving significant gains in visual fidelity and sequence scalability. Empirical comparisons show that, relative to concurrent models such as ~\cite{cui2024hallo3, yi2025magicinfinite, ominihuman, wang2025fantasytalking}, our proposed SkyReels-Audio system yields extended-duration videos with enhanced resolution and perceptual quality.

\section{Conclusion}

This paper presents SkyReels-Audio, a unified and omni framework for generating talking portraits conditioned exclusively on audio inputs. Our method leverages a hybrid training paradigm that aligns auditory and visual modalities, enabling precise modeling of the correlations between speech signals and corresponding lip articulations, facial expressions, and bodily gestures. To support the generation of videos of arbitrary length, we incorporate a dynamic sliding-window mechanism that ensures seamless temporal continuity and perceptual coherence across frames.
Extensive experimental evaluations—spanning both qualitative assessments and quantitative benchmarks—demonstrate that SkyReels-Audio consistently achieves superior performance in audio-visual synchronization and animation fidelity across a wide range of speaker identities, vocal characteristics, and multimodal conditioning scenarios.

\section{Contributors}
We gratefully acknowledge all contributors for their dedicated efforts. The following lists recognize participants by their primary contribution roles:
\begin{itemize}
    \item \textbf{Project Sponsor:} Yahui Zhou

    \item \textbf{Project Leaders:} Di Qiu, Guibin Chen

    \item \textbf{Core Contributors:} Zhengcong Fei, Hao Jiang, Baoxuan Gu, Youqiang Zhang, Jiahua Wang, Jialin Bai, Chaofeng Ao, Debang Li, Mingyuan Fan

\end{itemize}

\bibliographystyle{ieee}
\bibliography{main}

\end{document}